%% file: CameraReady2027.tex
\documentclass[letterpaper]{article} 
\usepackage{aaai2027}  
\usepackage[hyphens]{url}  
\usepackage{graphicx} 
\usepackage{natbib}  
\usepackage{caption} 
\usepackage{algorithm}
\usepackage{algorithmic}

\usepackage{newfloat}
\usepackage{listings}
\DeclareCaptionStyle{ruled}{labelfont=normalfont,labelsep=colon,strut=off} 
\floatstyle{ruled}
\newfloat{listing}{tb}{lst}{}
\floatname{listing}{Listing}

\usepackage{booktabs}

\usepackage{newfloat}
\usepackage{listings}
\usepackage[table]{xcolor}
\usepackage{booktabs}
\usepackage{adjustbox}
\definecolor{panelblue}{RGB}{232,241,250}
\definecolor{panelpurple}{RGB}{241,236,249}

\DeclareCaptionStyle{ruled}{labelfont=normalfont,labelsep=colon,strut=off} 
\floatstyle{ruled}
\newfloat{listing}{tb}{lst}{}
\floatname{listing}{Listing}

\usepackage{booktabs}
\newcommand{\modelsub}[1]{$_{\text{\scriptsize #1}}$}

\usepackage[hyphens]{url}  
\usepackage{graphicx} 
\usepackage{natbib}  
\usepackage{caption} 
\usepackage{algorithm}
\usepackage{algorithmic}
\usepackage{multirow}
\usepackage{etoc}
\usepackage{subcaption}
\usepackage{amsfonts} 
\usepackage{amsmath}
\usepackage{newfloat}
\usepackage{listings}
\usepackage[table]{xcolor}
\usepackage{booktabs}
\usepackage{adjustbox}
\definecolor{panelblue}{RGB}{232,241,250}
\definecolor{panelpurple}{RGB}{241,236,249}

\DeclareCaptionStyle{ruled}{labelfont=normalfont,labelsep=colon,strut=off} 
\floatstyle{ruled}
\newfloat{listing}{tb}{lst}{}
\floatname{listing}{Listing}

\nocopyright

\title{iBrain: A Unified Foundation Model Reading the Brain from Surface to Spikes}
\author{
    Ying Chen, Tiou Wang, Zhifeng Yue\corresponding
}

\affiliations{
    Chinese Institute for Brain Research\\

}

\begin{document}

\maketitle

\input{Section/1-abstract}
\input{Section/2-introduction}

\input{Section/3-related}

\input{Section/4-method}

\input{Section/5-result}
\input{Section/6-conclusion}

\bibliography{aaai2027}

\end{document}

%% file: Section/1-abstract.tex
\begin{abstract}
\begin{quote}

Invasive neural recordings provide high-fidelity measurements of brain activity, with signals such as intracranial EEG (iEEG) and intracortical spiking activity capturing neural dynamics at different spatial and temporal scales.
Yet existing neural foundation models have largely been developed independently for different invasive recording paradigms, leaving joint pretraining across heterogeneous invasive signals underexplored.
In this work, we introduce iBrain, a unified foundation model that jointly learns from iEEG and spiking activity.
iBrain employs signal-specific encoders to accommodate their distinct signal characteristics and a shared spatiotemporal Transformer backbone to model dependencies across recording channels and time.
We pretrain iBrain on over 7,000 hours of heterogeneous neural recordings using masked signal reconstruction and channel-view alignment, promoting contextual modeling of neural dynamics and robustness across different channels.
iBrain consistently outperforms single-signal pretraining baselines and achieves state-of-the-art performance on multiple benchmarks.
Further experiments demonstrate that iBrain exhibits transferability and data efficiency across diverse recording settings.
These results highlight the potential of joint pretraining on heterogeneous invasive neural recordings to support scalable neural modeling and transferable representations across recording settings and downstream tasks.
\end{quote}
\end{abstract}

%% file: Section/2-introduction.tex
\section{Introduction}

Neural recordings provide information-rich measurements of brain activity, allowing computational models to characterize behaviorally relevant neural dynamics and translate them into functional outputs~\cite{buzsaki2012origin}.
Such models have enabled advances in applications including speech neuroprostheses~\cite{moses2021neuroprosthesis,willett2023high}, movement decoding~\cite{collinger2013high}, and seizure monitoring~\cite{proix2021forecasting}.
Among existing recording paradigms, invasive approaches measure neural activity closer to its physiological sources, typically offering higher signal-to-noise ratios and more spatially localized access to task-relevant neural signals than non-invasive techniques~\cite{engel2005invasive,pesaran2018investigating}.
These advantages have made invasive neural recordings an important foundation for high-performance brain--computer interfaces~\cite{willett2021high,card2026long,wilson2025long}.

\begin{figure}[t]
\centering
\begin{subfigure}{\linewidth}
    \centering
    \includegraphics[width=\linewidth]{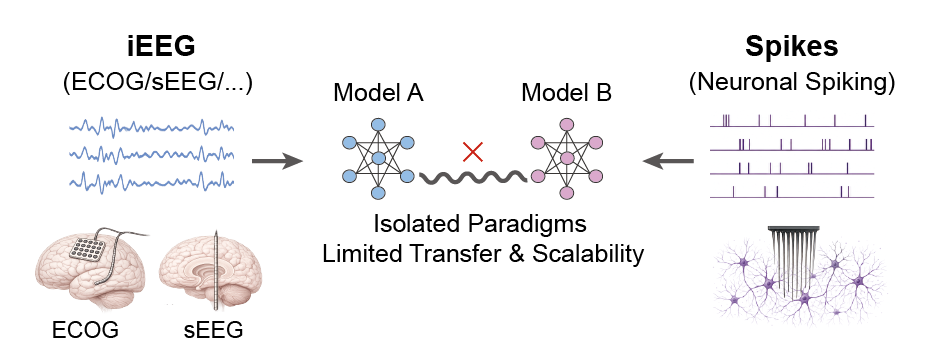}
    \caption{Previous methods}
\end{subfigure}
\vspace{0.5em}
\begin{subfigure}{\linewidth}
    \centering
    \includegraphics[width=\linewidth]{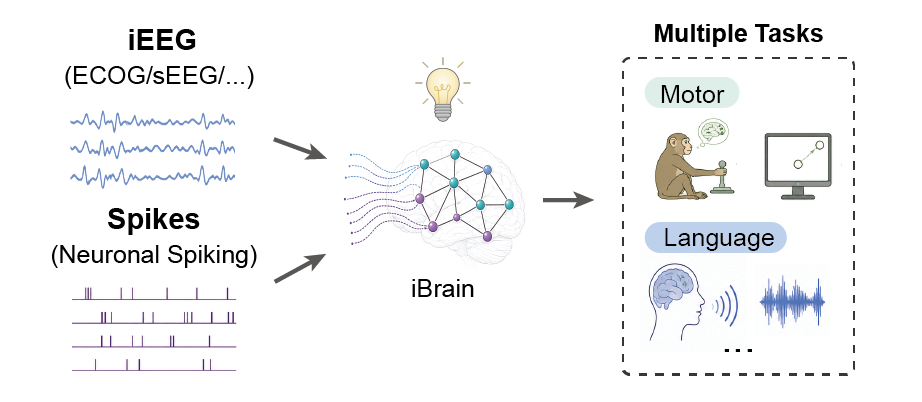}
    \caption{Our method}
\end{subfigure}
\caption{
\textbf{Comparison with previous methods. }
\textbf{a,} Previous methods often train dedicated models for individual invasive recording paradigms, resulting in limited transferability and scalability.
\textbf{b,} iBrain jointly learns from iEEG and spikes within a unified model, producing general neural representations for diverse downstream tasks.
}
\label{fig:comparison}
\end{figure}

Large-scale pretraining has emerged as an effective paradigm for learning transferable representations from complex temporal signals~\cite{zhang2024self,ma2024survey}.
The rich neural dynamics captured by invasive recordings make them particularly well suited to such representation learning~\cite{pesaran2018investigating}.
Recent studies have achieved encouraging results with pretrained foundation models for invasive neural recordings~\cite{wang2023brainbert,ye2023neural,azabou2025multi}, capturing shared neural dynamics across channels and time. %
These models improve downstream neural decoding, data-efficient adaptation, and cross-context generalization.


Despite these advances, existing methods remain limited in their utilization of heterogeneous invasive neural data. 
Intracranial EEG (iEEG), such as surface ECoG and
stereotactic EEG (sEEG), records extracellular field activity
across diverse cortical and deep-brain locations
~\cite{buzsaki2012origin,wu2024review}.
At a finer spatial scale, intracortical microelectrode recordings provide
access to spiking activity from localized neuronal populations
~\cite{williams2025vivo}.
These signals differ in form and observation scale, providing complementary measurements of underlying neural dynamics~\cite{einevoll2013modelling,buzsaki2012origin}. 
This raises a key question: \textbf{\textit{could joint learning across such heterogeneous invasive signals lead to better neural representations?}} 
As illustrated in Figure~\ref{fig:comparison}, we investigate this question by moving beyond separate models for individual recording types toward a unified framework for iEEG and spiking activity.
A positive answer would suggest that scarce and fragmented invasive recordings may be transformed into a foundation for learning more comprehensive and generalizable brain representations across scales.

In this work, we introduce iBrain, the first foundation model that
jointly learns from heterogeneous invasive neural recordings, from
cortical-surface and deep-brain iEEG to intracortical spiking activity.
iBrain integrates signal-specific encoders with a shared spatiotemporal attention backbone, projecting heterogeneous neural signals into a common representation space while preserving signal-specific information.
We pretrain iBrain with two self-supervised objectives: masked reconstruction to learn contextual signal representations, and channel-view alignment to improve robustness to channel variation.
For large-scale pretraining, we curate a heterogeneous corpus comprising 3,950 hours of iEEG and 3,209 hours of spiking activity.
We evaluate the resulting model on eight downstream benchmarks spanning motor decoding and language-related tasks.

To sum up, the main contributions of our work are as follows:

\begin{itemize}
    \item \textbf{Joint iEEG and spike learning.}
    We propose to jointly learn from iEEG and spiking activity, integrating complementary neural dynamics within a shared representation space.

    \item \textbf{Heterogeneous-signal and channel-view pretraining.}
    We introduce iBrain, the first foundation model to jointly learn from iEEG and spikes, using modality-specific reconstruction and channel-view alignment to learn robust representations across heterogeneous invasive neural recordings.

    \item \textbf{Large-scale heterogeneous corpus and diverse evaluation.}
    We curate a large-scale heterogeneous invasive neural corpus of over 7,000 hours for pretraining and evaluate iBrain on diverse downstream benchmarks covering motor decoding and language-related tasks.
\end{itemize}

%% file: Section/3-related.tex
\section{Related Work}

\subsection{Foundation Models for Invasive Neural Signals}

Large-scale self-supervised pretraining has emerged as a powerful paradigm for learning generalizable representations from complex data~\citep{awais2025foundation,liu2021self,krishnan2022self}.
Recent studies have begun to adapt foundation models to invasive neural recordings.
For iEEG recordings, BrainBERT~\citep{wang2023brainbert} learns reusable Transformer representations from unlabeled intracranial recordings through masked self-supervised pretraining, improving data-efficient neural decoding across subjects and tasks.
Brant~\citep{zhang2023brant} scales this direction by learning general-purpose representations from a large corpus of intracranial recordings and jointly modeling temporal and spectral information to support downstream tasks including neural signal forecasting, imputation, and seizure detection.
For spiking activity, the Neural Data Transformer family~\citep{ye2021representation,ye2023neural,ye2026generalist} extends Transformer-based pretraining to neural population activity and intracortical motor decoding.
POYO~\citep{azabou2023unified} introduces a unified tokenization and latent modeling framework for neural population decoding across sessions and animals.
Building on this framework, POYO+~\citep{azabou2025multi} supports multi-session and multi-task neural decoding across diverse cell types and brain regions, demonstrating transferable decoding on large-scale calcium-imaging recordings.
UniBCI~\citep{hong2026unibci} extends neural pretraining to heterogeneous invasive BCI datasets spanning multiple species, subjects, brain regions, and behavioral paradigms.
Together, these studies demonstrate the potential of foundation models for invasive neural recordings.
However, existing efforts have largely focused on individual signal types, leaving the diversity of invasive neural recordings underutilized for unified representation learning.

\subsection{Modeling Heterogeneous Neural Recordings}

Neural recordings exhibit substantial heterogeneity across acquisition principles, spatial layouts and physiological content, making unified representation learning challenging~\cite{hong2019novel,biessmann2011analysis,marblestone2013physical}. 
Recent studies have explored unified modeling frameworks for integrating heterogeneous neural modalities. 
Brain-OF~\citep{guo2026brain} models heterogeneity across fMRI, EEG, and MEG by mapping signals with different spatiotemporal resolutions into a shared semantic space, while combining modality-invariant and modality-specific experts for joint multimodal pretraining.
BrainOmni~\citep{xiao2026brainomni} develops a unified EEG--MEG pretraining framework with sensor-aware tokenization to capture variations in sensor layout, orientation, and type.
Recent work further bridged scalp EEG and intracranial EEG by incorporating pretrained neural representations and geometric constraints, aiming to recover high-fidelity intracranial-like information from scalp recordings~\citep{dong2026bridging}. 
While these studies highlight the value of unified modeling for heterogeneous brain recordings, they mainly focus on non-invasive modalities or scalp--intracranial alignment. How to jointly model heterogeneous invasive signals with distinct signal forms and observation scales remains underexplored.

%% file: Section/4-method.tex
\section{Method}

\begin{figure*}[h]
\centering
\includegraphics[width=\linewidth]{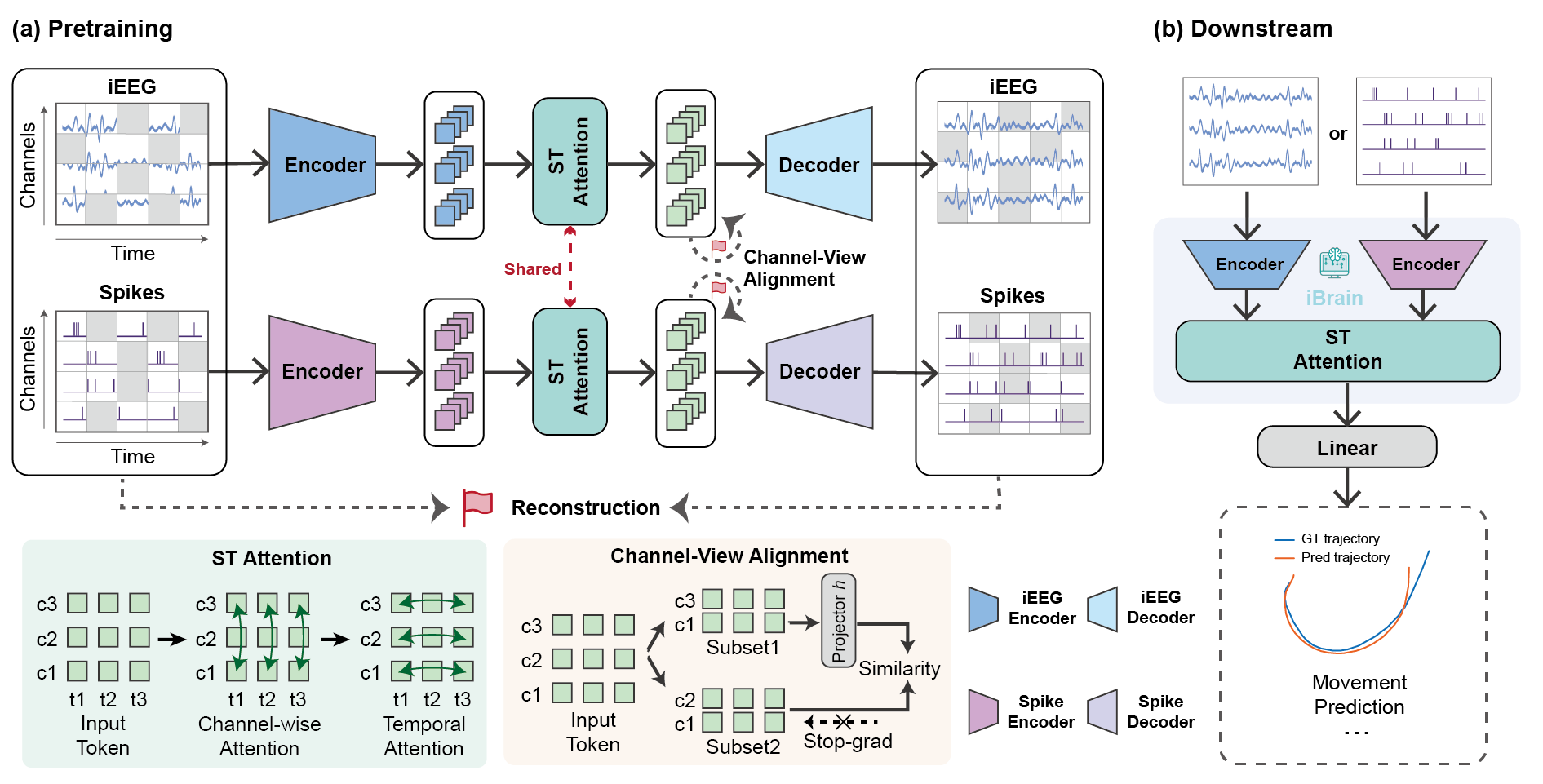}
\caption{
\textbf{Overview of iBrain.}
\textbf{(a) Pretraining.}
Signal-specific encoders transform masked iEEG waveform patches and binned
spike counts into channel–time tokens. A shared spatiotemporal (ST) attention backbone models dependencies across channels
and time, while signal-specific decoders reconstruct the masked signal content. Alongside masked reconstruction, channel-view alignment encourages consistent representations across different channel subsets. The bottom panels
illustrate the ST attention mechanism and channel-view alignment objective.
\textbf{(b) Downstream adaptation.}
The pretrained
signal-specific encoder and shared ST backbone are adapted to each downstream task with a lightweight task-specific head.
}
\label{fig:method}
\end{figure*}

We present iBrain, a unified foundation model for heterogeneous invasive
neural recordings spanning iEEG and spiking activity.
Figure~\ref{fig:method} provides an overview of the proposed method.
Signal-specific encoders map iEEG waveform patches and binned spike counts
into channel--time tokens.
The resulting tokens are jointly contextualized across channels and time
within a shared spatiotemporal attention backbone.
Pretraining jointly optimizes signal-specific masked reconstruction for iEEG
and spiking inputs and representation alignment across different channel
subsets.
For downstream evaluation, the corresponding signal-specific encoder and the
shared pretrained backbone are adapted to motor and language-related neural
decoding tasks using lightweight task-specific heads.
The following sections describe the three core components of iBrain:
signal-specific tokenization and encoding, shared spatiotemporal attention,
and self-supervised pretraining objectives.

\subsection{Signal-specific Tokenization and Encoding}

For a one-second segment of signal type
$m \in \{\mathrm{iEEG}, \mathrm{spike}\}$,
we divide each recording sequence into $S=10$ non-overlapping temporal
patches of 100~ms.
A recording sequence corresponds to an iEEG channel for
$m=\mathrm{iEEG}$ and to the binned count sequence of a single unit for
$m=\mathrm{spike}$.
The resulting patches are arranged as
$\mathbf{X}^{m} \in \mathbb{R}^{C_m \times S \times P_m}$,
where $C_m$ denotes the number of iEEG channels or spike units and $P_m$
denotes the number of waveform samples or count bins in each patch.
For iEEG sampled at 500~Hz, each patch contains
$P_{\mathrm{iEEG}}=50$ waveform samples.
For spiking activity represented using 20-ms bins, each patch contains
$P_{\mathrm{spike}}=5$ spike counts.
When recordings are padded along the recording dimension for batching, a
binary validity mask is used to exclude padded positions from attention,
pooling, and loss computation.

Because iEEG waveforms and spike counts have different input structures, we
use a separate patch encoder for each signal type.
Let
$\mathbf{x}^{m}_{c,s} \in \mathbb{R}^{P_m}$
denote the patch at recording index $c$ and temporal index $s$.
The corresponding encoder
$f_m:\mathbb{R}^{P_m}\rightarrow\mathbb{R}^{d}$
maps each patch independently to a $d$-dimensional token:
\begin{equation}
    \mathbf{h}^{m}_{c,s}
    =
    f_m\left(\mathbf{x}^{m}_{c,s}\right).
\end{equation}
For iEEG, $f_{\mathrm{iEEG}}$ is a temporal convolutional adapter with a
residual linear pathway.
The convolutional pathway extracts local waveform patterns within each
patch, while the residual pathway directly projects the waveform samples
into the token space.
For spike data, $f_{\mathrm{spike}}$ is an MLP adapter that encodes the
binned counts within each patch.
Applying the corresponding encoder to all patches produces
$\mathbf{H}^{m} \in \mathbb{R}^{C_m \times S \times d}$.

\subsection{Shared Spatiotemporal Attention}

iBrain employs a shared spatiotemporal (ST) attention backbone to model
channel-wise and temporal dependencies in the token embeddings produced by
the signal-specific encoders.
For each signal type
$m \in \{\mathrm{iEEG}, \mathrm{spike}\}$,
we augment the token embeddings with a learnable signal-type embedding and a
learnable temporal position embedding:
\begin{equation}
    \mathbf{Z}^{m}_{c,s}
    =
    \mathbf{H}^{m}_{c,s}
    +
    \mathbf{e}^{m}_{\mathrm{type}}
    +
    \mathbf{e}_{s}^{\mathrm{time}},
\end{equation}
where $\mathbf{e}^{m}_{\mathrm{type}}$ identifies the signal type and
$\mathbf{e}_{s}^{\mathrm{time}}$ encodes the temporal patch index.

The backbone consists of stacked criss-cross ST attention blocks.
Each block factorizes attention over the channel--time grid into successive
channel and temporal attention.
Channel attention is first applied independently at each temporal index to
capture dependencies across iEEG channels or spike units:
\begin{equation}
    \widetilde{\mathbf{Z}}^{m}_{:,s}
    =
    \mathrm{Attn}_{\mathrm{ch}}
    \left(
        \mathbf{Z}^{m}_{:,s}
    \right),
    \qquad s=1,\ldots,S.
\end{equation}
Temporal attention is subsequently applied along the temporal dimension of
each channel or unit:
\begin{equation}
    \overline{\mathbf{Z}}^{m}_{c,:}
    =
    \mathrm{Attn}_{\mathrm{time}}
    \left(
        \widetilde{\mathbf{Z}}^{m}_{c,:}
    \right),
    \qquad c=1,\ldots,C_m.
\end{equation}
A learnable relative temporal bias is added to the temporal attention logits
to encode the relative distances between temporal patches.
Each block further contains pre-normalization, residual connections, and a
feed-forward network.
Denoting the complete ST backbone by $F_{\mathrm{ST}}$, its output is
\begin{equation}
    \mathbf{U}^{m}
    =
    F_{\mathrm{ST}}\left(\mathbf{Z}^{m}\right)
    \in
    \mathbb{R}^{C_m \times S \times d}.
\end{equation}

Together, the channel and temporal attention stages model dependencies along
the recording and temporal dimensions.
Sharing the ST backbone across iEEG and spike inputs imposes a shared
spatiotemporal inductive bias, encouraging the model to capture channel--time
structures rather than representations specific to either signal type.

\subsection{Self-supervised Pretraining Objectives}

\paragraph{Masked reconstruction.}
Given an input segment
$\mathbf{X}^{m} \in \mathbb{R}^{C_m \times S \times P_m}$,
we sample a binary mask
$\mathbf{M} \in \{0,1\}^{C_m \times S}$
over the channel--time grid, where
$\mathbf{M}_{c,s}=1$ indicates that the entire patch
$\mathbf{x}^{m}_{c,s}$ is masked.
Masking is performed in the original signal space before patch encoding:
\begin{equation}
    \widetilde{\mathbf{x}}^{m}_{c,s}
    =
    \left(1-\mathbf{M}_{c,s}\right)
    \mathbf{x}^{m}_{c,s}.
\end{equation}
The masked patches are then processed by the corresponding signal-specific
encoder.
Following the embedding and backbone computations described above, this
produces contextualized representations
$\widetilde{\mathbf{U}}^{m}
\in
\mathbb{R}^{C_m \times S \times d}$.

A signal-specific two-layer MLP decoder
$g_m:\mathbb{R}^{d}\rightarrow\mathbb{R}^{P_m}$
maps each contextualized token back to its original patch space:
\begin{equation}
    \widehat{\mathbf{x}}^{m}_{c,s}
    =
    g_m\left(\widetilde{\mathbf{U}}^{m}_{c,s}\right)
    =
    \mathbf{W}^{m}_{2}
    \phi\left(
        \mathbf{W}^{m}_{1}
        \widetilde{\mathbf{U}}^{m}_{c,s}
        +
        \mathbf{b}^{m}_{1}
    \right)
    +
    \mathbf{b}^{m}_{2},
\end{equation}
where $\phi(\cdot)$ denotes the nonlinear activation.
The patch-level predictions form
$\widehat{\mathbf{X}}^{m}
\in
\mathbb{R}^{C_m \times S \times P_m}$.
Reconstruction losses are evaluated only at masked positions corresponding
to valid channels or units.

For iEEG, the target is the channel-normalized waveform patch.
The per-patch reconstruction error is
\begin{equation}
    \ell_{\mathrm{iEEG}}
    \left(
        \widehat{\mathbf{x}},
        \mathbf{x}
    \right)
    =
    \frac{1}{P_{\mathrm{iEEG}}}
    \left\|
        \widehat{\mathbf{x}}
        -
        \mathbf{x}
    \right\|_2^2.
\end{equation}
The iEEG reconstruction loss is therefore
\begin{equation}
    \mathcal{L}_{\mathrm{rec}}^{\mathrm{iEEG}}
    =
    \frac{
        \sum_{c,s}
        \mathbf{V}_{c}
        \mathbf{M}_{c,s}
        \,
        \ell_{\mathrm{iEEG}}
        \left(
            \widehat{\mathbf{x}}^{\mathrm{iEEG}}_{c,s},
            \mathbf{x}^{\mathrm{iEEG}}_{c,s}
        \right)
    }{
        \sum_{c,s}
        \mathbf{V}_{c}\mathbf{M}_{c,s}
    },
\end{equation}
where $\mathbf{V}_{c}$ indicates whether recording index $c$ corresponds to
a valid channel rather than padding.

For spike data, the decoder predictions are transformed into positive
Poisson rates:
\begin{equation}
    \boldsymbol{\lambda}_{c,s}
    =
    \mathrm{softplus}
    \left(
        \widehat{\mathbf{x}}^{\mathrm{spike}}_{c,s}
    \right)
    +
    \epsilon,
\end{equation}
where $\epsilon$ is a small constant for numerical stability.
The spike reconstruction loss is
\begin{equation}
    \mathcal{L}_{\mathrm{rec}}^{\mathrm{spike}}
    =
    \frac{
        \sum_{c,s}
        \mathbf{V}_{c}
        \mathbf{M}_{c,s}
        \,
        \mathrm{PoissonNLL}
        \left(
            \boldsymbol{\lambda}_{c,s},
            \mathbf{x}^{\mathrm{spike}}_{c,s}
        \right)
    }{
        \sum_{c,s}
        \mathbf{V}_{c}\mathbf{M}_{c,s}
    }.
\end{equation}
The two reconstruction objectives reflect the different statistical
properties of iEEG waveforms and spike counts.

\paragraph{Channel-view alignment.}
In addition to masked reconstruction, we introduce channel-view alignment to
improve robustness to variable and partially observed recording
configurations.
For each segment of signal type $m$, we sample two overlapping subsets of its
valid iEEG channels or spike units, producing two channel views of the same
segment.
Each view is processed by the signal-specific encoder and shared ST backbone,
and the valid channel--time tokens are globally pooled to obtain
segment-level representations
$\mathbf{r}^{m}_{1}$ and $\mathbf{r}^{m}_{2}$.
A projection head maps these representations to
$\mathbf{q}^{m}_{1}$ and $\mathbf{q}^{m}_{2}$, and a prediction head produces
$\mathbf{p}^{m}_{1}$ and $\mathbf{p}^{m}_{2}$.
Following SimSiam~\cite{chen2021exploring}, the alignment loss is
\begin{equation}
    \mathcal{L}_{\mathrm{align}}^{m}
    =
    \frac{1}{2}
    D\left(
        \mathbf{p}^{m}_{1},
        \mathrm{sg}\left(\mathbf{q}^{m}_{2}\right)
    \right)
    +
    \frac{1}{2}
    D\left(
        \mathbf{p}^{m}_{2},
        \mathrm{sg}\left(\mathbf{q}^{m}_{1}\right)
    \right),
\end{equation}
where $\mathrm{sg}(\cdot)$ denotes stop-gradient and
$D(\cdot,\cdot)$ denotes negative cosine similarity.
This objective encourages the segment-level representation to remain stable
under channel or unit subsampling.

\paragraph{Pretraining schedule.}
To train the shared backbone across the two corpora, we alternate between
iEEG and spike minibatches at a 1:1 step ratio, following a shared-parameter
training schedule for unpaired multimodal data~\cite{gupta2025better}.
Each optimization step contains a minibatch from a single signal type.
Let
$m_t \in \{\mathrm{iEEG},\mathrm{spike}\}$
denote the signal type sampled at step $t$.
The corresponding encoder and reconstruction decoder are used together with
the shared ST backbone, and the objective for that step is
\begin{equation}
    \mathcal{L}_{t}
    =
    \mathcal{L}_{\mathrm{rec}}^{m_t}
    +
    \mathcal{L}_{\mathrm{align}}^{m_t}.
\end{equation}
The signal-specific modules are therefore optimized using their corresponding
inputs, while the shared ST backbone receives updates from both signal types
throughout pretraining.

%% file: Section/5-result.tex
\section{Experiments}

\subsection{Datasets}

\noindent \textbf{Pretraining Datasets.}
We curate a large-scale corpus of invasive neural recordings for self-supervised pretraining, spanning diverse species, recording interfaces, brain regions, and behavioral contexts.
The iEEG corpus mainly consists of human ECoG and sEEG recordings from AJILE12~\cite{peterson2022ajile12} and the SWEC iEEG Dataset~\cite{carzaniga2025foundation}, while the spike corpus is from Neural Pile~\cite{orhan2025neural}. 
All recordings are divided into non-overlapping one-second windows; iEEG signals are resampled to 500 Hz, and spike recordings are represented as binned spike-count sequences over the same duration. 
The resulting corpus contains 4,662,027 AJILE12 samples, 9,560,582 SWEC samples, and 11,553,750 Neural Pile samples, corresponding to 3,950 hours of iEEG recordings and 3,209 hours of spiking activity.

\noindent \textbf{Downstream Datasets.} 
We evaluate iBrain on eight downstream benchmarks covering spike-based movement decoding, as well as iEEG language perception. 
The spike benchmarks consist of MC-Maze~\cite{pei2021neural}, Area2-Bump~\cite{pei2021neural}, Perich T-CO~\cite{perich2025long}, and Perich T-RT~\cite{perich2025long}, spanning delayed maze-constrained reaching, perturbation-driven somatosensory responses, center-out reaching, and random-target reaching. 
The iEEG benchmarks are constructed from Brain Treebank~\cite{wang2024brain} and evaluate neural responses to audio volume, pitch, sentence onset, and speech/non-speech events during naturalistic movie viewing.



















\subsection{Implementation and Settings}

\noindent\textbf{Pretraining details.}
iBrain uses a 6-layer Transformer backbone with a hidden dimension of 256,
8 attention heads, a feed-forward hidden dimension of 1024, and a dropout
rate of 0.1.
For masked reconstruction, 50\% of the valid tokens are randomly masked and
reconstructed using MLP decoders.
At each training step, two channel-subset views are sampled from the same
recording, each retaining approximately 80\% of the valid channels, with
75\% overlap and at least two channels.
Their pooled representations are aligned using a cosine objective in a
128-dimensional projection space.
We pretrain iBrain for 30 epochs using AdamW with a learning rate of
$5\times10^{-4}$, a weight decay of $5\times10^{-2}$, gradient clipping at
1.0, and a learning-rate schedule comprising 2,000 linear warm-up steps
followed by cosine annealing to a minimum learning rate of
$1\times10^{-5}$.
Pretraining is performed on 8 GPUs with a per-GPU batch size of 32 for each
signal type.

\noindent \textbf{Evaluation Strategy.} 
For downstream evaluation, we adapt iBrain by fine-tuning all pretrained parameters together with lightweight task-specific prediction heads. 
For spike-based benchmarks, including MC-Maze, Area2-Bump, Perich T-CO, and Perich T-RT, each dataset is split into 80\% training data and 20\% test data. 
For the four Brain Treebank tasks, we follow the subject-specific evaluation protocol of BrainBERT~\cite{wang2023brainbert} and PopT~\cite{ICLR2025_acb3e200}, where the model is fine-tuned on a subset of recordings from each subject and evaluated on the remaining held-out recordings.

\noindent \textbf{Baselines \& Metrics.} 
We compare iBrain with representative baselines tailored to each signal family. 
For iEEG benchmarks, we include Brant~\cite{zhang2023brant}, BrainBERT~\cite{wang2023brainbert}, and TOTEM~\cite{talukder2024totem}, where BrainBERT and TOTEM are evaluated with linear and PopT~\cite{ICLR2025_acb3e200} prediction heads. 
For spike-based benchmarks, we compare against NDT1~\cite{ye2021representation}, NDT2~\cite{ye2023neural}, MtM~\cite{zhang2024towards}, POYO~\cite{azabou2023unified}, and UniBCI~\cite{hong2026unibci}. 
We report area under the ROC curve (AUC) for the Brain TreeBank classification tasks and the coefficient of determination ($R^2$) for spike-based regression benchmarks, with higher values indicating better performance.

\begin{table}[h]
\centering
\setlength{\tabcolsep}{5pt}
\renewcommand{\arraystretch}{1.08}
\begin{adjustbox}{max width=\linewidth}
\begin{tabular}{lcccc}
\toprule
\rowcolor{panelblue}
\multicolumn{5}{c}{\textbf{iEEG benchmark}} \\
\rowcolor{panelblue}\textbf{Method} 
& \textbf{Pitch} 
& \textbf{Volume} 
& \textbf{Onset} 
& \textbf{Speech} \\
\midrule
Brant 
& 0.60 & 0.75 & 0.81 & 0.79 \\
BrainBERT\modelsub{+Linear}  
& 0.57 & 0.64 & 0.72 & 0.72 \\
BrainBERT\modelsub{+PopT}  
& 0.55 & 0.62 & 0.72 & 0.68 \\
TOTEM\modelsub{+Linear} 
& 0.58 & 0.66 & 0.78 & 0.79 \\ 
TOTEM\modelsub{+PopT} 
& 0.52 & 0.64 & 0.81 & 0.75 \\
\textbf{iBrain} 
& \textbf{0.68} & \textbf{0.89} & \textbf{0.86} & \textbf{0.89} \\
\midrule[0.6pt]

\rowcolor{panelpurple}
\multicolumn{5}{c}{\textbf{Spike benchmark}} \\
\rowcolor{panelpurple}\textbf{Method}
& \textbf{MC-Maze}
& \textbf{Area2-Bump}
& \textbf{T-CO}
& \textbf{T-RT} \\
\midrule
NDT1
& 0.868 & 0.886 & 0.749 & 0.674 \\
NDT2
& 0.899 & 0.872 & 0.724 & 0.696 \\
MtM
& 0.863 & 0.890 & 0.694 & 0.608 \\
POYO
& 0.887 & 0.866 & 0.767 & 0.678 \\
UniBCI
& 0.876 & 0.894 & 0.742 & \textbf{0.716} \\
\textbf{iBrain}
& \textbf{0.914} & \textbf{0.903} & \textbf{0.785} & 0.692 \\
\bottomrule
\end{tabular}
\end{adjustbox}
\caption{
Comparison with representative baseline methods on downstream benchmarks.
}
\label{tab:baselines}
\end{table}

\begin{table*}[h]
\centering
\setlength{\tabcolsep}{5.5pt}
\renewcommand{\arraystretch}{1.10}
\begin{tabular}{lcccccccc}
\toprule
\textbf{Dataset}
& \textbf{Pitch}
& \textbf{Volume}
& \textbf{Onset}
& \textbf{Speech}
& \textbf{MC-Maze}
& \textbf{Area2-Bump}
& \textbf{Perich T-CO}
& \textbf{Perich T-RT} \\
\midrule
iEEG only
& 0.68 & 0.85 & 0.84 & 0.86
& -- & -- & -- & -- \\

Spike only
& -- & -- & -- & --
& 0.89 & 0.89 & 0.77 & 0.64 \\

iEEG + Spike
& 0.68 & 0.89 & 0.86 & 0.89
& 0.91 & 0.90 & 0.79 & 0.69 \\
\bottomrule
\end{tabular}
\caption{Comparison of iBrain under iEEG-only, spike-only, and joint iEEG--spike pretraining.}
\label{tab:pretrain_modality_ablation}
\end{table*}
\subsection{Comparison with Baselines}
We evaluate iBrain against representative baselines on downstream benchmarks, with results summarized in Table~\ref{tab:baselines}. On the Brain TreeBank iEEG tasks, iBrain consistently outperforms all baselines across the four AUC-based classification tasks. Compared with the strongest baseline on each task, iBrain improves Pitch from 0.60 to 0.68, Volume from 0.75 to 0.89, Onset from 0.81 to 0.86, and Speech from 0.79 to 0.89. These results suggest that heterogeneous pretraining can provide useful representations for iEEG language tasks.

On spike-based benchmarks, iBrain also achieves competitive performance across diverse motor and perturbation-related decoding tasks. It obtains the best results on MC-Maze, Area2-Bump, and Perich T-CO, improving over the strongest baseline. 
On Perich T-RT, iBrain does not achieve the best result, suggesting that further improvements are needed for some random-target reaching settings.
Overall, iBrain obtains the top performance on seven of the eight evaluated benchmarks.

\subsection{Joint Pretraining}

\begin{figure}[h]
\centering
\includegraphics[width=\linewidth]{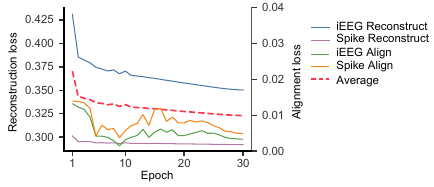}
\caption{Joint pretraining loss curves.}
\label{fig:loss}
\end{figure}

We first examine the training dynamics of joint pretraining. As shown in Figure~\ref{fig:loss}, the total loss decreases steadily over epochs, indicating stable convergence when iEEG and spike trains are trained together. The iEEG reconstruction loss shows a clear downward trend, while the spike reconstruction loss remains low and stable. In addition, both alignment losses stay small throughout training, suggesting that channel-view consistency regularizes the shared representation without interfering with signal-specific reconstruction.

We then examine whether joint pretraining improves downstream transfer. As shown in Table~\ref{tab:pretrain_modality_ablation}, joint iEEG~+~Spike pretraining achieves the best overall results across both task groups, improving over iEEG-only pretraining on Speech and Volume, and over spike-only pretraining on MC-Maze, Area2-Bump, Perich T-CO, and Perich T-RT. These results suggest that heterogeneous pretraining preserves signal-specific strengths while producing a more general representation space for both language-related and motor decoding tasks.

\subsection{Few-shot Evaluation}

\begin{figure}[h]
\includegraphics[width=\linewidth]{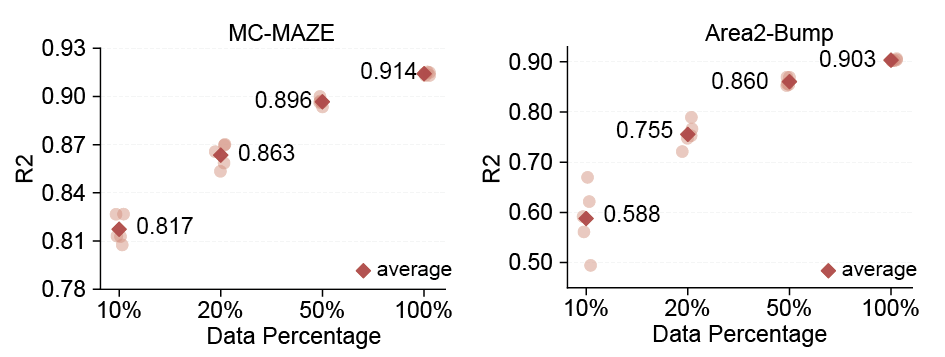}
\caption{Few-shot neural decoding with limited labeled trials.
Average $R^2$ on MC-Maze and Area2-Bump when fine-tuning iBrain with different fractions of labeled training trials.}
\label{fig:fewshot}
\end{figure}

To evaluate the label efficiency of iBrain, we conduct few-shot experiments on MC-Maze and Area2-Bump by fine-tuning the pretrained model with 10\%, 20\%, 50\%, and 100\% of the labeled training trials. As shown in Figure~\ref{fig:fewshot}, iBrain maintains competitive performance under limited supervision and improves consistently as more labeled data become available. On MC-Maze, the average $R^2$ increases from 0.817 with 10\% labels to 0.863, 0.896, and 0.914 with 20\%, 50\%, and 100\% labels, respectively. On Area2-Bump, the average $R^2$ improves from 0.588 to 0.755, 0.860, and 0.903 across the same label fractions. The improvements are particularly pronounced in the low-label regime, especially on Area2-Bump, where increasing the labeled fraction from 10\% to 20\% yields an absolute gain of 0.167 in $R^2$. Notably, using only 50\% of the labeled trials results in performance gaps of only 0.018 and 0.043 relative to full-data training on MC-Maze and Area2-Bump, respectively. These results suggest that pretraining provides an effective initialization for downstream adaptation and reduces the amount of task-specific supervision required for neural decoding.

\begin{table}[h]
\centering
\small
\setlength{\tabcolsep}{6pt}
\renewcommand{\arraystretch}{1.08}
\begin{tabular}{lcccc}
\toprule
\multirow{2}{*}{\textbf{Model}} 
& \multicolumn{2}{c}{\textbf{MC-Maze}} 
& \multicolumn{2}{c}{\textbf{Area2-Bump}} \\
\cmidrule(lr){2-3} \cmidrule(lr){4-5}
& Fine-tune & Frozen
& Fine-tune & Frozen \\
\midrule
Full 
& \textbf{0.914} & \textbf{0.904} 
& \textbf{0.903} & \textbf{0.867} \\
w/o channel align. 
& 0.893 & 0.872 
& 0.886 & 0.855 \\
w/o channel attn. 
& 0.889 & 0.881 
& 0.875 & 0.842 \\
w/o time attn. 
& 0.901 & 0.902 
& 0.874 & 0.850 \\
\bottomrule
\end{tabular}
\caption{
Model ablation results on MC-Maze and Area2-Bump. We report mean $R^2$ under both fine-tuning and frozen-backbone settings.
}
\label{tab:model_ablation}
\end{table}

\subsection{Model Ablation}

We further conduct ablation studies to evaluate the contributions of the
channel-view alignment objective and the spatiotemporal attention design.
As shown in Table~\ref{tab:model_ablation}, the full model achieves the
highest mean $R^2$ on both MC-Maze and Area2-Bump under the fine-tuning and
frozen-backbone settings.
Removing the channel-view alignment objective reduces performance across all
settings, with the largest decrease observed on MC-Maze with a frozen
backbone, where $R^2$ drops from 0.904 to 0.872.
This result suggests that channel-view alignment improves the transferability
of the pretrained representations under frozen-backbone evaluation.
Removing either channel or temporal attention also reduces performance,
supporting the importance of modeling both cross-channel interactions and
temporal dependencies.
In particular, removing channel attention consistently degrades performance
across both benchmarks and evaluation settings, whereas removing temporal
attention has a larger effect on Area2-Bump than on MC-Maze.
These different degradation patterns suggest that channel and temporal
attention provide contributions to downstream decoding.

\subsection{Scaling Analysis}
\begin{figure}[h]
\includegraphics[width=\linewidth]{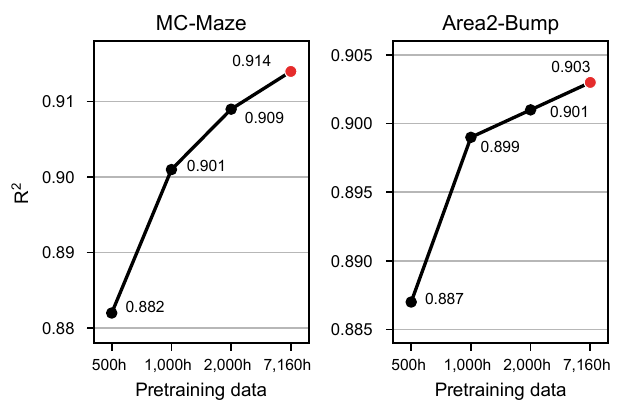}
\caption{Effect of pretraining data scale.
Downstream $R^2$ on MC-Maze and Area2-Bump for models pretrained on 500, 1,000, 2,000, and 7,160 hours of neural recordings. Equal durations of iEEG and spike recordings are included
at each scale.
}
\label{fig:Scaling_curve}
\end{figure}

We study the effect of pretraining data scale using 500, 1,000, 2,000, and 7,160 hours of neural recordings, with equal amounts of iEEG and spike data at each scale. As shown in Figure~\ref{fig:Scaling_curve}, downstream performance improves monotonically with increasing pretraining data, with $R^2$ increasing from 0.882 to 0.914 on MC-Maze and from 0.887 to 0.903 on Area2-Bump. The magnitude of improvement is relatively modest, and the gains become smaller beyond 2,000 hours. This scaling behavior is less pronounced than that often observed in general-domain foundation models, suggesting that simply increasing recording duration may not be sufficient for neural representation learning. One possible explanation is that neural recordings contain substantial redundancy within subjects and sessions, such that the effective diversity of the data may increase more slowly than the total number of recording hours. These observations suggest that future scaling may benefit from expanding subject, task, and acquisition diversity in addition to increasing the overall data volume.

\subsection{Encoder Representations}

\begin{figure}[h]
\includegraphics[width=\linewidth]{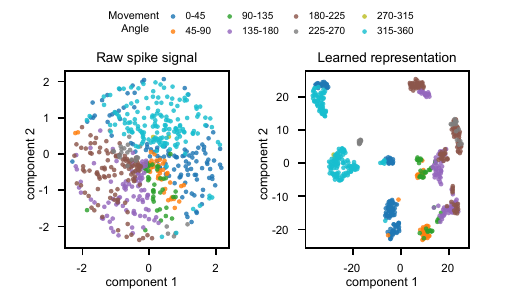}
\caption{t-SNE visualization on MC-Maze.
Raw spike counts are highly overlapping, whereas learned representations show clearer separation across movement directions.}
\label{fig:t-SNE}
\end{figure}

To qualitatively examine the structure of the learned neural
representations, we perform t-SNE visualization~\cite{van2008visualizing} on the held-out test split
of MC-Maze.
MC-Maze contains macaque motor-cortical spike recordings collected during
center-out reaching, and we group the trials into eight movement-angle ranges
for visualization.
As shown in Figure~\ref{fig:t-SNE}, the raw spike-count signals are highly
entangled in the low-dimensional space, with substantial overlap among
samples from different movement directions.
In contrast, representations extracted from the fine-tuned iBrain encoder
exhibit a more organized distribution.
Samples with similar movement angles tend to cluster locally, while groups
corresponding to different angle ranges are more clearly separated.
This visualization suggests that fine-tuning transforms sparse spike-count
sequences into representations that better reflect movement-related
structure on held-out trials.

%% file: Section/6-conclusion.tex
\section{Conclusion}
In this paper, we introduced iBrain, a unified foundation model for heterogeneous invasive neural recordings that jointly learns from cortical-surface and deep-brain iEEG as well as intracortical spiking activity. By integrating signal-specific encoders and reconstruction objectives with a shared spatiotemporal Transformer and channel-view alignment, iBrain accommodates distinct signal characteristics while learning representations that remain effective across diverse recording settings. Pretrained on more than 7,000 hours of neural recordings, iBrain achieves the strongest performance on seven of eight downstream benchmarks spanning motor decoding and language-related tasks, while the ablation results confirm the complementary contributions of spatiotemporal modeling and channel-view alignment. Joint pretraining further improves upon signal-specific pretraining, and the few-shot and scaling analyses demonstrate improved label efficiency and consistent benefits from increasing pretraining data, while also suggesting that future gains may depend on greater diversity across subjects, tasks, and acquisition settings rather than recording duration alone. Taken together, these results support joint pretraining across heterogeneous invasive neural signals as a practical and scalable direction toward more transferable neural representations across recording interfaces and downstream tasks.

%% file: aaai2027.bib
@article{hong2026unibci,
  title={UniBCI: Towards a Unified Pretrained Model for Invasive Brain-Computer Interfaces},
  author={Hong, Binjie and Xiong, Rui and Han, Liyuan and Zhang, Tielin},
  journal={arXiv preprint arXiv:2605.00061},
  year={2026}
}

@article{guo2026brain,
  title={Brain-OF: An Omnifunctional Foundation Model for fMRI, EEG and MEG},
  author={Guo, Hanning and Abdellatif, Farah and Bi, Hanwen and Galbenus, Andrei and Shah, Jon and Morrison, Abigail and Dammers, J{\"u}rgen and others},
  journal={arXiv preprint arXiv:2602.23410},
  year={2026}
}

@article{wang2023brainbert,
  title={BrainBERT: Self-supervised representation learning for intracranial recordings},
  author={Wang, Christopher and Subramaniam, Vighnesh and Yaari, Adam Uri and Kreiman, Gabriel and Katz, Boris and Cases, Ignacio and Barbu, Andrei},
  journal={arXiv preprint arXiv:2302.14367},
  year={2023}
}

@article{zhang2023brant,
  title={Brant: Foundation model for intracranial neural signal},
  author={Zhang, Daoze and Yuan, Zhizhang and Yang, Yang and Chen, Junru and Wang, Jingjing and Li, Yafeng},
  journal={Advances in Neural Information Processing Systems},
  volume={36},
  pages={26304--26321},
  year={2023}
}

@article{awais2025foundation,
  title={Foundation models defining a new era in vision: a survey and outlook},
  author={Awais, Muhammad and Naseer, Muzammal and Khan, Salman and Anwer, Rao Muhammad and Cholakkal, Hisham and Shah, Mubarak and Yang, Ming-Hsuan and Khan, Fahad Shahbaz},
  journal={IEEE Transactions on Pattern Analysis and Machine Intelligence},
  volume={47},
  number={4},
  pages={2245--2264},
  year={2025},
  publisher={IEEE}
}

@article{ye2021representation,
  title={Representation learning for neural population activity with neural data transformers},
  author={Ye, Joel and Pandarinath, Chethan},
  journal={arXiv preprint arXiv:2108.01210},
  year={2021}
}

@article{ye2023neural,
  title={Neural data transformer 2: multi-context pretraining for neural spiking activity},
  author={Ye, Joel and Collinger, Jennifer and Wehbe, Leila and Gaunt, Robert},
  journal={Advances in Neural Information Processing Systems},
  volume={36},
  pages={80352--80374},
  year={2023}
}

@article{ye2026generalist,
  title={A generalist intracortical motor decoder},
  author={Ye, Joel and Rizzoglio, Fabio and Ma, Xuan and Smoulder, Adam and Mao, Hongwei and Blumenthal, Gary and Hockeimer, William and Kunigk, Nicolas and Moore, Dalton and Marino, Patrick and others},
  journal={Advances in Neural Information Processing Systems},
  volume={38},
  pages={110547--110587},
  year={2026}
}

@article{azabou2023unified,
  title={A unified, scalable framework for neural population decoding},
  author={Azabou, Mehdi and Arora, Vinam and Ganesh, Venkataramana and Mao, Ximeng and Nachimuthu, Santosh and Mendelson, Michael and Richards, Blake and Perich, Matthew and Lajoie, Guillaume and Dyer, Eva},
  journal={Advances in Neural Information Processing Systems},
  volume={36},
  pages={44937--44956},
  year={2023}
}

@inproceedings{azabou2025multi,
  title={Multi-session, multi-task neural decoding from distinct cell-types and brain regions},
  author={Azabou, Mehdi and Pan, Krystal and Arora, Vinam and Knight, Ian and Dyer, Eva and Richards, Blake A},
  booktitle={International Conference on Learning Representations},
  volume={2025},
  pages={59654--59677},
  year={2025}
}

@article{xiao2026brainomni,
  title={Brainomni: A brain foundation model for unified eeg and meg signals},
  author={Xiao, Qinfan and Cui, Ziyun and Zhang, Chi and Chen, Siqi and Wu, Wen and Thwaites, Andrew and Woolgar, Alexandra and Zhou, Bowen and Zhang, Chao},
  journal={Advances in Neural Information Processing Systems},
  volume={38},
  pages={41179--41212},
  year={2026}
}

@article{dong2026bridging,
  title={Bridging scalp and intracranial EEG in BCI via pretrained neural representations and geometric constraint embedding},
  author={Dong, Yihang and Jing, Changhong and Wang, Shuqiang},
  journal={arXiv preprint arXiv:2604.14202},
  year={2026}
}

@article{peterson2022ajile12,
  title={Ajile12: Long-term naturalistic human intracranial neural recordings and pose},
  author={Peterson, Steven M and Singh, Satpreet H and Dichter, Benjamin and Scheid, Michael and Rao, Rajesh PN and Brunton, Bingni W},
  journal={Scientific data},
  volume={9},
  number={1},
  pages={184},
  year={2022},
  publisher={Nature Publishing Group UK London}
}

@article{carzaniga2025foundation,
  title={A foundation model with multi-variate parallel attention to generate neuronal activity},
  author={Carzaniga, Francesco and Hersche, Michael and Sebastian, Abu and Schindler, Kaspar and Rahimi, Abbas},
  journal={arXiv preprint arXiv:2506.20354},
  year={2025}
}

@inproceedings{orhan2025neural,
  title={The Neural Pile: 476 billion tokens of broad-coverage spiking neural activity data},
  author={Orhan, Emin and Wang, Feiyi},
  booktitle={NeurIPS 2025 Workshop on Foundation Models for the Brain and Body},
  year= {2025}
}

@article{perich2025long,
  title={Long-term recordings of motor and premotor cortical spiking activity during reaching in monkeys},
  author={Perich, Matthew G and Miller, Lee E and Azabou, Mehdi and Dyer, Eva L},
  journal={Data set},
  year={2025}
}

@article{buzsaki2012origin,
  title={The origin of extracellular fields and currents—EEG, ECoG, LFP and spikes},
  author={Buzs{\'a}ki, Gy{\"o}rgy and Anastassiou, Costas A and Koch, Christof},
  journal={Nature reviews neuroscience},
  volume={13},
  number={6},
  pages={407--420},
  year={2012},
  publisher={Nature Publishing Group UK London}
}

@article{willett2023high,
  title={A high-performance speech neuroprosthesis},
  author={Willett, Francis R and Kunz, Erin M and Fan, Chaofei and Avansino, Donald T and Wilson, Guy H and Choi, Eun Young and Kamdar, Foram and Glasser, Matthew F and Hochberg, Leigh R and Druckmann, Shaul and others},
  journal={Nature},
  volume={620},
  number={7976},
  pages={1031--1036},
  year={2023},
  publisher={Nature Publishing Group UK London}
}

@article{collinger2013high,
  title={High-performance neuroprosthetic control by an individual with tetraplegia},
  author={Collinger, Jennifer L and Wodlinger, Brian and Downey, John E and Wang, Wei and Tyler-Kabara, Elizabeth C and Weber, Douglas J and McMorland, Angus JC and Velliste, Meel and Boninger, Michael L and Schwartz, Andrew B},
  journal={The Lancet},
  volume={381},
  number={9866},
  pages={557--564},
  year={2013},
  publisher={Elsevier}
}

@article{proix2021forecasting,
  title={Forecasting seizure risk in adults with focal epilepsy: a development and validation study},
  author={Proix, Timoth{\'e}e and Truccolo, Wilson and Leguia, Marc G and Tcheng, Thomas K and King-Stephens, David and Rao, Vikram R and Baud, Maxime O},
  journal={The Lancet Neurology},
  volume={20},
  number={2},
  pages={127--135},
  year={2021},
  publisher={Elsevier}
}

@article{engel2005invasive,
  title={Invasive recordings from the human brain: clinical insights and beyond},
  author={Engel, Andreas K and Moll, Christian KE and Fried, Itzhak and Ojemann, George A},
  journal={Nature Reviews Neuroscience},
  volume={6},
  number={1},
  pages={35--47},
  year={2005},
  publisher={Nature Publishing Group UK London}
}

@article{willett2021high,
  title={High-performance brain-to-text communication via handwriting},
  author={Willett, Francis R and Avansino, Donald T and Hochberg, Leigh R and Henderson, Jaimie M and Shenoy, Krishna V},
  journal={Nature},
  volume={593},
  number={7858},
  pages={249--254},
  year={2021},
  publisher={Nature Publishing Group UK London}
}

@article{card2026long,
  title={Long-term independent use of an intracortical brain--computer interface for speech and cursor control},
  author={Card, Nicholas S and Singer-Clark, Tyler and Peracha, Hamza and Iacobacci, Carrina and Hou, Xianda and Wairagkar, Maitreyee and Fogg, Zachery and Offenberg, Elena C and Hochberg, Leigh R and Stavisky, Sergey D and others},
  journal={Nature Medicine},
  pages={1--7},
  year={2026},
  publisher={Nature Publishing Group}
}

@article{wilson2025long,
  title={Long-term unsupervised recalibration of cursor-based intracortical brain--computer interfaces using a hidden Markov model},
  author={Wilson, Guy H and Stein, Elias A and Kamdar, Foram and Avansino, Donald T and Pun, Tsam Kiu and Gross, Ronnie and Hosman, Tommy and Singer-Clark, Tyler and Kapitonava, Anastasia and Hochberg, Leigh R and others},
  journal={Nature Biomedical Engineering},
  pages={1--19},
  year={2025},
  publisher={Nature Publishing Group UK London}
}

@article{zhang2024self,
  title={Self-supervised learning for time series analysis: Taxonomy, progress, and prospects},
  author={Zhang, Kexin and Wen, Qingsong and Zhang, Chaoli and Cai, Rongyao and Jin, Ming and Liu, Yong and Zhang, James Y and Liang, Yuxuan and Pang, Guansong and Song, Dongjin and others},
  journal={IEEE transactions on pattern analysis and machine intelligence},
  volume={46},
  number={10},
  pages={6775--6794},
  year={2024},
  publisher={IEEE}
}

@article{ma2024survey,
  title={A survey on time-series pre-trained models},
  author={Ma, Qianli and Liu, Zhen and Zheng, Zhenjing and Huang, Ziyang and Zhu, Siying and Yu, Zhongzhong and Kwok, James T},
  journal={IEEE Transactions on Knowledge and Data Engineering},
  volume={36},
  number={12},
  pages={7536--7555},
  year={2024},
  publisher={IEEE}
}

@article{pesaran2018investigating,
  title={Investigating large-scale brain dynamics using field potential recordings: analysis and interpretation},
  author={Pesaran, Bijan and Vinck, Martin and Einevoll, Gaute T and Sirota, Anton and Fries, Pascal and Siegel, Markus and Truccolo, Wilson and Schroeder, Charles E and Srinivasan, Ramesh},
  journal={Nature neuroscience},
  volume={21},
  number={7},
  pages={903--919},
  year={2018},
  publisher={Nature Publishing Group US New York}
}

@article{williams2025vivo,
  title={In vivo microelectrode arrays for neuroscience},
  author={Williams, Nathaniel P and Voroslakos, Mihaly and Shi, Delin and Pwint, May Yoon and Lanzio, Vittorino and Mao, Hongwei and Zolotavin, Pavlo and Yoon, Euisik and Stieglitz, Thomas and Xie, Chong and others},
  journal={Nature Reviews Methods Primers},
  volume={5},
  number={1},
  pages={31},
  year={2025},
  publisher={Nature Publishing Group UK London}
}

@inproceedings{chen2021exploring,
  title={Exploring simple siamese representation learning},
  author={Chen, Xinlei and He, Kaiming},
  booktitle={Proceedings of the IEEE/CVF conference on computer vision and pattern recognition},
  pages={15750--15758},
  year={2021}
}

@article{pei2021neural,
  title={Neural latents benchmark'21: evaluating latent variable models of neural population activity},
  author={Pei, Felix and Ye, Joel and Zoltowski, David and Wu, Anqi and Chowdhury, Raeed H and Sohn, Hansem and O'Doherty, Joseph E and Shenoy, Krishna V and Kaufman, Matthew T and Churchland, Mark and others},
  journal={arXiv preprint arXiv:2109.04463},
  year={2021}
}

@article{wang2024brain,
  title={Brain treebank: Large-scale intracranial recordings from naturalistic language stimuli},
  author={Wang, Christopher and Yaari, Adam and Singh, Aaditya K and Subramaniam, Vighnesh and Rosenfarb, Dana and DeWitt, Jan and Misra, Pranav and Madsen, Joseph R and Stone, Scellig and Kreiman, Gabriel and others},
  journal={Advances in Neural Information Processing Systems},
  volume={37},
  pages={96505--96540},
  year={2024}
}

@inproceedings{ICLR2025_acb3e200,
 author = {Chau, Geeling and Wang, Christopher and Talukder, Sabera and Subramaniam, Vighnesh and Soedarmadji, Saraswati and Yue, Yisong and Katz, Boris and Barbu, Andrei},
 booktitle = {International Conference on Learning Representations},
 editor = {Y. Yue and A. Garg and N. Peng and F. Sha and R. Yu},
 pages = {69100--69121},
 title = {Population Transformer: Learning Population-level Representations of Neural Activity},
 url = {https://proceedings.iclr.cc/paper_files/paper/2025/file/acb3e20075b0a2dfa3565f06681578e5-Paper-Conference.pdf},
 volume = {2025},
 year = {2025}
}

@article{
talukder2024totem,
title={{TOTEM}: {TO}kenized Time Series {EM}beddings for General Time Series Analysis},
author={Sabera J Talukder and Yisong Yue and Georgia Gkioxari},
journal={Transactions on Machine Learning Research},
issn={2835-8856},
year={2024},
url={https://openreview.net/forum?id=QlTLkH6xRC},
note={}
}

@article{zhang2024towards,
  title={Towards a" universal translator" for neural dynamics at single-cell, single-spike resolution},
  author={Zhang, Yizi and Wang, Yanchen and Jim{\'e}nez-Benet{\'o}, Donato M and Wang, Zixuan and Azabou, Mehdi and Richards, Blake and Tung, Renee and Winter, Olivier and Dyer, Eva and Paninski, Liam and others},
  journal={Advances in Neural Information Processing Systems},
  volume={37},
  pages={80495--80521},
  year={2024}
}

@article{moses2021neuroprosthesis,
  title={Neuroprosthesis for decoding speech in a paralyzed person with anarthria},
  author={Moses, David A and Metzger, Sean L and Liu, Jessie R and Anumanchipalli, Gopala K and Makin, Joseph G and Sun, Pengfei F and Chartier, Josh and Dougherty, Maximilian E and Liu, Patricia M and Abrams, Gary M and others},
  journal={New England Journal of Medicine},
  volume={385},
  number={3},
  pages={217--227},
  year={2021},
  publisher={Mass Medical Soc}
}

@article{wu2024review,
  title={A review of motor brain-computer interfaces using intracranial electroencephalography based on surface electrodes and depth electrodes},
  author={Wu, Xiaolong and Metcalfe, Benjamin and He, Shenghong and Tan, Huiling and Zhang, Dingguo},
  journal={IEEE Transactions on Neural Systems and Rehabilitation Engineering},
  volume={32},
  pages={2408--2431},
  year={2024},
  publisher={IEEE}
}

@article{liu2021self,
  title={Self-supervised learning: Generative or contrastive},
  author={Liu, Xiao and Zhang, Fanjin and Hou, Zhenyu and Mian, Li and Wang, Zhaoyu and Zhang, Jing and Tang, Jie},
  journal={IEEE transactions on knowledge and data engineering},
  volume={35},
  number={1},
  pages={857--876},
  year={2021},
  publisher={IEEE}
}

@article{krishnan2022self,
  title={Self-supervised learning in medicine and healthcare},
  author={Krishnan, Rayan and Rajpurkar, Pranav and Topol, Eric J},
  journal={Nature Biomedical Engineering},
  volume={6},
  number={12},
  pages={1346--1352},
  year={2022},
  publisher={Nature Publishing Group UK London}
}

@article{biessmann2011analysis,
  title={Analysis of multimodal neuroimaging data},
  author={Biessmann, Felix and Plis, Sergey and Meinecke, Frank C and Eichele, Tom and Muller, Klaus-Robert},
  journal={IEEE reviews in biomedical engineering},
  volume={4},
  pages={26--58},
  year={2011},
  publisher={IEEE}
}

@article{marblestone2013physical,
  title={Physical principles for scalable neural recording},
  author={Marblestone, Adam H and Zamft, Bradley M and Maguire, Yael G and Shapiro, Mikhail G and Cybulski, Thaddeus R and Glaser, Joshua I and Amodei, Dario and Stranges, P Benjamin and Kalhor, Reza and Dalrymple, David A and others},
  journal={Frontiers in computational neuroscience},
  volume={7},
  pages={137},
  year={2013},
  publisher={Frontiers Media SA}
}

@article{hong2019novel,
  title={Novel electrode technologies for neural recordings},
  author={Hong, Guosong and Lieber, Charles M},
  journal={Nature Reviews Neuroscience},
  volume={20},
  number={6},
  pages={330--345},
  year={2019},
  publisher={Nature Publishing Group UK London}
}

@article{einevoll2013modelling,
  title={Modelling and analysis of local field potentials for studying the function of cortical circuits},
  author={Einevoll, Gaute T and Kayser, Christoph and Logothetis, Nikos K and Panzeri, Stefano},
  journal={Nature Reviews Neuroscience},
  volume={14},
  number={11},
  pages={770--785},
  year={2013},
  publisher={Nature Publishing Group UK London}
}

@article{gupta2025better,
  title={Better together: Leveraging unpaired multimodal data for stronger unimodal models},
  author={Gupta, Sharut and Sundaram, Shobhita and Wang, Chenyu and Jegelka, Stefanie and Isola, Phillip},
  journal={arXiv preprint arXiv:2510.08492},
  year={2025}
}

@article{van2008visualizing,
  title={Visualizing data using t-SNE.},
  author={Van der Maaten, Laurens and Hinton, Geoffrey},
  journal={Journal of machine learning research},
  volume={9},
  number={11},
  year={2008}
}
